%% file: manuscript_arXiv.tex
\documentclass[conference]{IEEEtran}
\IEEEoverridecommandlockouts
\usepackage{cite}
\usepackage{amsmath,amssymb,amsfonts}
\usepackage{graphicx}
\usepackage{textcomp}
\usepackage{xcolor}
\usepackage{subcaption}
\usepackage{booktabs,multirow,tabularx}
\usepackage{wrapfig}
\usepackage{url}
\usepackage[hidelinks]{hyperref} 
\usepackage{xr}
\usepackage{xspace} 
\title{Back on Track: Deployment-time Policy Improvement via World Model Planning}

\author{Colton Crosby$^{1*}$, Hao Wang$^{1, 2*}$, Mac Schwager$^{1}$ and Somil Bansal$^{1}$
\thanks{*The authors contributed equally to this work.}
\thanks{$^{1}$The authors are associated with the Aeronautics and Astronautics Department, Stanford University.
        {\tt\small \{cjcrosby, haowwang, schwager, somil\}@stanford.edu}}%
\thanks{$^{2}$The author is associated with the Ming Hsieh Department of Electrical and Computer Engineering, University of Southern California.
      }%
\thanks{$\dagger$This research is supported in part by the Stanford Robotics Center and the NSF CAREER Program under award 224016.}
}

\begin{document}

\input{macros}
\IEEEaftertitletext{
 \centering
 \vspace{-2em}
 \includegraphics[width=0.98\textwidth]{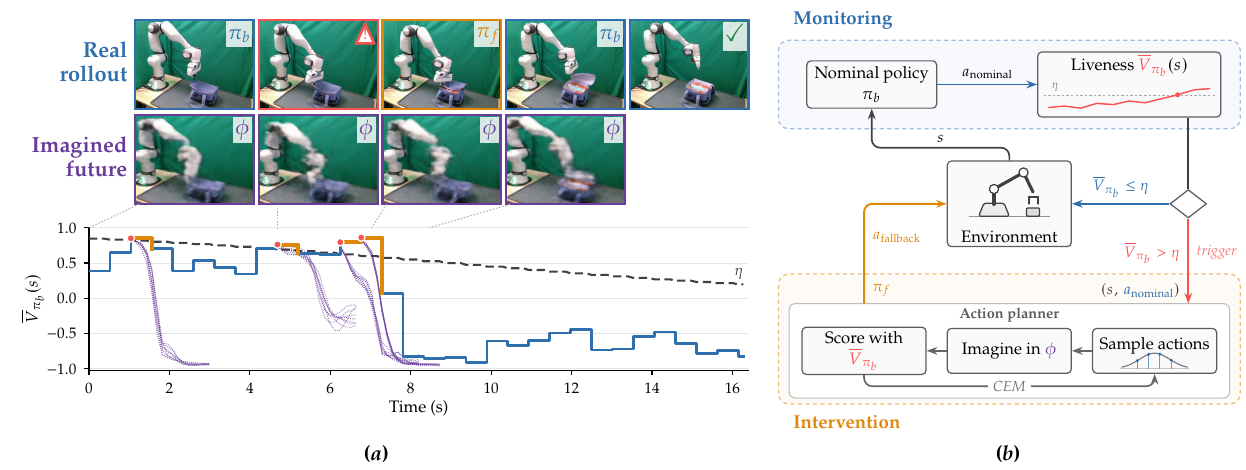}
 \captionof{figure}{We address deployment-time improvement of visuomotor manipulation policies, without retraining or otherwise modifying the nominal policy; assuming only black-box access. We cast policy improvement in a monitoring-and-intervention framework \textbf{(\textit{b})}: a learned value function $\vfmean$ continuously monitors the nominal policy $\pibase$ during execution which, upon threshold exceedance, triggers an action planner to intervene by rolling out candidate futures in a learned world model ($\wm$), optimizing actions against $\vfmean$ as its planning objective. A real-time rollout \textbf{(\textit{a})} demonstrates $\vfmean$ exceeding a threshold $\eta$, alternating control between the planner $\pif$ and the nominal policy $\pibase$ and imagining futures (purple traces; a lower value means closer to task success) that optimize the value function to bring $\pi_b$ back into states from which it can succeed.}
 \label{fig:header}
}

\maketitle

\begin{abstract}
State-of-the-art visuomotor policies are increasingly capable of executing complex tasks, yet can fail during deployment in ways they cannot predict or recover from on their own. Improving their performance and robustness typically requires collecting more data, retraining, or modifying the policy architecture, all of which are time- and resource-intensive in practice. 
We instead propose an offline-trained monitoring-and-intervention framework that improves a policy during deployment using only black-box query access. From a limited set of labeled on-policy rollouts, we learn a policy-conditioned liveness value that measures the nominal policy's progress toward task completion; we additionally train a compact latent world model using these rollouts together with play data. At deployment, the liveness value continuously monitors execution. When it detects a loss of progress, a short-horizon planner searches the world model for corrective actions using the same value as its planning objective, after which control returns to the nominal policy. We evaluate our method on the LIBERO-Goal task suite and on a Franka Panda manipulator, using the vision-language-action model $\pi_{0.5}$ and a diffusion policy as nominal policies, and demonstrate significant improvements in task success rates, while supporting real-time control.
Additional results and videos are available on the \href{https://policy-improvement-wm.pages.dev/}{project website}.

\end{abstract}

\section{Introduction}
Visuomotor policies are increasingly capable of complex tasks from rich sensory observations~\cite{open_vla, gemini_robotics_1_5, pi06_star}, yet they remain brittle at deployment and can fail in ways they cannot recover from on their own.

 The dominant response is to scale policy learning itself, with more data, more compute, and larger architectures~\cite{oxe_dataset, droid_dataset}.

While effective, these approaches are expensive and require access to the policy's parameters and training pipeline --- access a practitioner increasingly does not have, since strong policies are increasingly served as inference APIs or shipped with closed weights. When the weights are out of reach, deployment-time correction is a promising avenue.

 We therefore ask \textit{how a policy that has already been trained and deployed can be improved without modifying the policy itself.} We assume only black-box access: the parameters, architecture, and action-generation procedure are unknown, and the policy can only be queried for actions. Fine-tuning and methods that steer the policy's internal sampling are therefore inapplicable, and improvement must come from an external wrapper that monitors execution and intervenes through actions alone. Two requirements make this demanding: policy rollouts are costly, so the wrapper must be learned from few labeled on-policy executions, and all monitoring and planning must keep pace with the control loop while the policy executes.

Our approach is based on monitoring-and-intervention, or \emph{filtering}~\cite{ames_2016_cbf_qp, borquez2024safety}, which in control theory most often enforces safety: a filter watches the nominal policy and overrides it only when a constraint would be violated~\cite{bansalHamiltonJacobiReachabilityBrief2017b, wang_2024_coop_safe_perf}. We instead filter for task completion. Rather than characterizing and avoiding every failure mode, an intervention seeks to return the system to a state from which the nominal policy can complete the task. The intuition is that a capable policy need not be replaced or retrained when it falters; a short corrective action is often enough to put it back in its basin of success.

 The framework has two components. A monitor tracks the nominal policy's execution and detects when it is no longer progressing toward task completion; we instantiate it with a policy-conditioned liveness value function~\cite{wang2026_policy_eval}, learned from on-policy rollouts with sparse episode-level outcome labels, that estimates how close the policy is to completing the task. When the value indicates a loss of progress, a planner briefly overrides the policy and searches a compact latent world model --- trained on the same rollouts plus auxiliary play data --- for a short corrective action sequence, then hands control back.

 Critically, the same value that triggers intervention also scores the futures the world model imagines, directing the planner toward states from which the nominal policy can recover.

 Methods that generate video futures report seconds-scale intervention times~\cite{xu2026vlaintheloop, fan2026dreamavoid}, forcing either a pause or a control rate too low for responsive manipulation. We therefore design around deployment latency: the world model is compact, candidate futures are evaluated in parallel, and planning emits an action chunk rather than a single action. In our hardware experiment a planning cycle takes $108$\ms, which, amortized over the chunk, keeps the monitor--planner loop above $10$\,Hz.

We summarize our contributions as follows. First, we formulate deployment-time improvement of visuomotor manipulation policies under a strict black-box access model and cast it as liveness filtering: an external wrapper intervenes briefly to return a fixed nominal policy to its basin of success. Second, we present a real-time instantiation in which a single policy-conditioned liveness value both monitors execution and guides corrective planning in a compact latent world model, with a data-informed action proposal that makes the search efficient. Third, we validate it across policy classes: on the ten-task LIBERO-Goal suite our method improves $\pi_{0.5}$~\cite{pi05} from $72.6\%$ to $81.5\%$, outperforming offline-RL and safety-filtering baselines, and on a Franka Panda manipulator it improves a diffusion policy~\cite{diffusion_policy} from $40\%$ to $60\%$ while operating within the policy's control loop.

\section{Related Work}

We group prior work by the access and supervision it assumes, noting in each case how our problem setting differs. One line of work intervenes inside the policy's action-generation process, requiring white-box access to the policy. Latent Policy Barrier~\cite{latent_policy_barrier} treats the latent embeddings of expert demonstrations as a barrier separating in- from out-of-distribution states, using a learned dynamics model to steer a diffusion policy's denoising noise back toward the expert distribution at inference, equating in-distribution behavior with task success. VLS~\cite{liu2026vls} steers a pretrained policy's denoising process with guidance from a vision-language model; and \cite{wagenmaker2025_steer_dp_latent_rl} replaces a diffusion policy's Gaussian sampling noise with a distribution learned via reinforcement learning (RL). 
These methods intervene directly in action generation, but rely on the policy's internal representations or sampling procedure and so do not apply when the nominal policy is only a black box.

A second class of methods treats the nominal policy as a black box and intervenes only through its inputs or outputs. RoboMonkey~\cite{kwok2025robomonkey} and Steer Your Generalists~\cite{nakamoto2024_steer_generalist} rerank action candidates sampled from the policy. This avoids modifying the policy's weights, but requires multiple policy queries at every decision point, which is prohibitive for real-time control: a single diffusion-policy query alone takes $180$\ms in our hardware setting. We instead query $\pibase$ once per intervention and obtain candidate diversity from our own proposal distribution rather than from the policy's sampler.

Most closely related to our work are methods that use a learned world model to improve a fixed policy at deployment. Latent Policy Steering~\cite{wang2025latent} continuously searches a latent world model for improved action sequences. In contrast, our monitor invokes planning selectively and returns control to the nominal policy after a short correction, limiting unnecessary computation and intervention when the nominal policy is progressing successfully. 

DreamAvoid~\cite{fan2026dreamavoid} ranks futures from a pretrained video world model at annotated critical phases, and VLA-in-the-Loop~\cite{xu2026vlaintheloop} specializes the paradigm to grasping, rolling back to an earlier state on a predicted grasp failure. Both are scoped to predefined critical events and to supervision we do not assume: access to the VLA's sampling procedure plus critical-phase, progress, and failure-boundary labels for the former, grasp-feasibility labels and state rollback for the latter. We assume only black-box access and learn a policy-conditioned liveness value from sparse episode-level outcomes; the same value both detects loss of progress and guides planning, so intervention is not confined to predefined events, such as the wrong-task behavior in Fig.~\ref{fig:libero_t0_viz}. Their video models are also slower at deployment, reporting $2$\,s (DreamAvoid) and $20$\,s (VLA-in-the-Loop) per intervention against our $108$\ms planning cycle.

Monitoring-and-intervention filters have a principled basis in reachability analysis~\cite{bansalHamiltonJacobiReachabilityBrief2017b}, which formalizes intervention under known dynamics: a filter overrides the policy only when needed to keep the system within a desired set of states. This machinery supports the dual objectives of \textit{safety} (avoiding a failure set) and \textit{liveness} (reaching a goal set~\cite{borquez2024safety}). We focus on the latter.

Classical liveness filtering computes the reachable set exactly via dynamic programming~\cite{borquez2024safety}, but this is intractable for high-dimensional visual observations and the contact-rich dynamics of manipulation. We therefore build on the learned policy-conditioned liveness value of~\cite{wang2026_policy_eval}, which can be estimated from limited policy-rollout data with sparse, episode-level outcome labels. Whereas that prior work uses the liveness value for offline policy evaluation, we use it here both to detect loss of progress during execution and, critically, to guide corrective planning in a learned world model.

World models differ in what they predict, and that choice sets the deployment cost. Video-generation models synthesize pixels, which is expensive enough to put them at seconds-scale latency; DreamerV3 predicts distributions over discrete latent codes~\cite{hafnerMasteringDiverseDomains2024}; feature-prediction models forecast the patch-level features of a pretrained visual encoder, as in DINO-WM~\cite{dino_wm}. We take the last approach and keep the model compact, since each intervention must evaluate many imagined action sequences within a real-time budget.

\section{Problem Formulation}
We consider the problem of \emph{deployment-time policy improvement}. Given an episodic robotic task $\task$ and a fixed visuomotor policy $\pibase$ trained to perform $\task$, our goal is to increase the task success rate of $\pibase$ during execution without modifying or retraining the policy itself. Our formulation applies to episodic tasks with a measurable completion outcome; in this work, we focus on visuomotor manipulation.

We assume black-box access to $\pibase$: we do not know its parameters, architecture, or internal representations. 
The only information available is a dataset $\mathcal{D} = \{(\state^{i}_{1:T_i},\, \action^{i}_{1:T_i-1},\, y^i)\}_{i=1}^{N}$ of $N$ rollouts generated by executing $\pibase$ on $\task$. Each rollout $i$ consists of a state-action trajectory of length $T_i$ and an episode-level binary outcome $y^i \in \{0, 1\}$ indicating whether the task was completed. These sparse outcome labels are the only task-specific supervision assumed by our method. To learn a dynamics model, we may additionally use an auxiliary dataset $\mathcal{D}_{\mathrm{play}}$ of state-action transitions, without requiring outcome labels or any specific generation policy.

We formulate deployment-time improvement as a \emph{monitoring-and-intervention} problem. During execution, the nominal policy $\pibase$ acts by default. An external monitor $\monitor$ observes the execution of $\pibase$ and determines whether intervention is necessary. When $\monitor$ triggers, a fallback policy $\pif$ temporarily overrides $\pibase$ and 
applies a short corrective intervention, after which control is returned to $\pibase$. Neither component accesses or modifies the internals of the nominal policy. Our objective is to design $\monitor$ and $\pif$ from $\mathcal{D}$ and optionally $\mathcal{D}_{\mathrm{play}}$ to maximize task completion probability within the latency budget of the deployment control loop.

\section{Method}

We now present a concrete instantiation of the monitoring-and-intervention framework. A shared, policy-conditioned liveness value couples its two components: a monitor $\monitor$, which detects when the nominal policy $\pibase$ has lost progress toward task completion, and a short-horizon planner, acting as the fallback policy $\pif$, which searches a learned world model $\wm$ for a corrective action sequence under that same value. On trigger, the planner overrides $\pibase$, executes a short correction, and returns control to the nominal policy.

\subsection{Monitor}\label{sec:monitor}
Let $o$ denote the image observation and $q$ the proprioceptive state provided to $\pibase$. We define the state as $\state = [\psi(o),\ q]$, where $\psi(\cdot)$ is an image encoder (e.g., SigLIP2~\cite{siglip2}).
As the monitor, we employ the \emph{liveness value function} $\vf$ of the nominal policy $\pibase$, introduced in~\cite{wang2026_policy_eval}. 
This value is learned from the policy-rollout dataset $\mathcal{D}$ using the Bellman update

\begin{equation} \label{eq:liveness_operator}
 \vf(\state) = (1-\gamma) + \gamma\min\{l(\state), \vf(\state')\}
\end{equation}

where $\state'$ is the next state, $\gamma$ is the discount factor, and $l(\state)$ is a sparse goal function that is $-1$ at a goal state and $+1$ otherwise. 
In practice, we construct $l$ from the episode-level outcome labels in $\mathcal{D}$: the final state of a successful episode is labeled as a goal, while all other states receive $+1$.
Unlike a standard value function that sums discounted reward along a trajectory, Eq.~\eqref{eq:liveness_operator} uses the $\min$ operator to propagate goal attainment backward through the nominal policy's rollouts. The resulting value $\vf$ is policy-conditioned: it characterizes progress toward task completion under the future behavior of $\pibase$, rather than progress under an arbitrary controller. It takes values in $[-1,1]$, falling toward $-1$ as $\pibase$ approaches completion and rising toward $+1$ otherwise, so the monitor triggers on \emph{high} values and the planner \emph{minimizes} $\vf$. 

 Because the learned value is noisy --- especially when $\mathcal{D}$ is small --- we build the monitor from an ensemble of value networks, whose mean prediction we denote $\vfmean(\state)$, together with temporal smoothing over a window of size $W$. 
Specifically, at time $t$, we average the current prediction with the previous $W-1$ predictions,
\begin{equation}\label{eq:v_ens}
\widetilde{V}(\state_t)
=
\frac{1}{W}
\sum_{j=0}^{W-1}
\vfmean(\state_{t-j}),
\end{equation}
using $W=3$ in our experiments. 
The monitor triggers whenever $\widetilde{V}(\state_t) > \threshold_t$, where $\threshold_t$ is a time-varying threshold. We decay $\threshold_t$ from $+1$ to a terminal value $\hat{\threshold}=0$ according to a sigmoid schedule. Concretely $\threshold_t = [1+\exp((t-t_c)/\sigma)]^{-1}$ with $t$ the frame index within the episode and $t_c\!=\!60$, $\sigma\!=\!10$, so $\threshold_t$ falls from $0.998$ at the first frame to below $0.01$ by frame $120$ to allow roughly three seconds for the manipulator to leave its initial configuration, during which the liveness value is high even on successful executions and an intervention would be a false alarm. The same schedule is used for all ten tasks. We ablate the ensemble, temporal smoothing, and threshold schedule in Tab.~\ref{tab:ablations}.

\subsection{Planner}

 The planner optimizes an action sequence over a horizon $\ph$ and executes only its first $\ch\leq\ph$ actions before returning control. This short horizon matches the fallback policy's role: it seeks a corrective intervention rather than attempting to complete the task itself.

\subsubsection{Dynamics Model}
 Planning must run inside the control loop, so we learn a world model $\wm$ over the compact state representation $\state$ rather than in pixel space: given a state--action pair $(\state_t,\action_t)$ it predicts the next state, $\imgstate_{t+1} = \wm(\state_t,\action_t)$. We parameterize $\wm$ as a $29.6$M-parameter, four-layer ViT with sinusoidal positional encodings~\cite{dino_wm}. This prioritizes inference latency over capacity: each planning cycle evaluates many imagined trajectories while the robot executes.

We train $\wm$ using two sources of data. The policy-rollout dataset $\mathcal{D}$ provides transitions from regions of the state-action space visited by $\pibase$. We supplement these transitions with auxiliary play data $\mathcal{D}_{\mathrm{play}}$, which broadens action coverage beyond the nominal policy's behavior, a common practice in world model training~\cite{learning_latent_plans_from_play,wm_play_chen2025spa}.

To improve multi-step prediction with a compact model, we train $\wm$ autoregressively under a rollout-length curriculum.
At curriculum length $\lengthcurr$ the model is applied to its own predictions from $\imgstate_t=\state_t$, minimizing $\mathcal{L}_{\mathrm{WM}} = \sum_{j=1}^{\lengthcurr}
\left|\imgstate_{t+j}-\state_{t+j}\right|_2^2$.
The curriculum gradually increases $\lengthcurr$ from $1$ to $16$, first learning accurate local transitions and then exposing the model to its own prediction errors over longer rollouts.

\subsubsection{Planning Objective}
The planning objective $\cost(\cdot)$ encodes the desired effect of an intervention: returning the system to states from which the nominal policy can complete the task. Given the current state $\state_0$ and a candidate action sequence $\action_{0:\ph-1}$, the world model generates the imagined trajectory $\imgstate_1=\wm(\state_0,\action_0)$ and $\imgstate_i=\wm(\imgstate_{i-1},\action_{i-1})$ for $i>1$.
We score the candidate by summing the ensemble-mean liveness value along this trajectory:

\vspace{-8pt}
\begin{equation} \label{eqn:cost_function}
 \cost(\action_{0:\ph-1}) = \sum_{i=1}^{\ph} \vfmean(\imgstate_i)
\end{equation}

The planner minimizes Eq.~\eqref{eqn:cost_function}. Using the cumulative value, rather than only the terminal prediction, favors action sequences that make sustained progress throughout the imagined rollout. Critically, $\vfmean$ is the same policy-conditioned value used by the monitor. The planner therefore does not seek to complete the task on its own; it searches for states that are favorable under the subsequent behavior of $\pibase$.

\subsubsection{Planning Algorithm}
We minimize Eq.~\eqref{eqn:cost_function} with the Cross-Entropy Method (CEM), a parallelizable sampling-based optimizer suited to the non-smooth, non-convex objective a learned value function induces. Candidates must be diverse enough to find a correction, yet implausible actions push the world model off its training distribution and yield exploitable prediction errors; we therefore sample from a data-informed proposal centered on the nominal policy's action.

Vanilla CEM perturbs each action coordinate independently with isotropic Gaussian noise, which for action chunks yields temporally incoherent candidates; iCEM~\cite{pinneri2021sample} instead samples colored noise with a $1/f^\beta$ spectrum, imposing smoothness as a generic prior. However, this prior is not informed by the temporal and cross-dimensional correlations present in task-relevant actions.

We instead \textit{estimate} these correlations directly from the rollout data. We flatten each training-time action chunk of horizon $\ph$ and action dimension $D$ into a vector in $\mathbb{R}^{\ph D}$ and compute its empirical covariance $\Sigma$. Let $V_r$ and $\Lambda_r$ contain the leading $r$ eigenvectors and eigenvalues of $\Sigma$. We sample an action perturbation as
\begin{equation}\label{eq:pca_noise}
 \delta = s\,V_k \Lambda_k^{1/2}\varepsilon,
 \qquad \varepsilon \sim \mathcal{N}(0, I_k),
\end{equation}

where we use the $k$ of the $HD$ eigenvectors that explain $98\%$ of the measured variance (e.g., $k\!\!=\!\!13$ for $H\!\!=\!\!10,D\!\!=\!\!6$ for our $\pi_{0.5}$ LIBERO case), and set the exploration scale to $s\!=\!0.6$. The gripper dimension is near-binary, so a continuous Gaussian subspace models it poorly; we leave it unperturbed and use the nominal policy's gripper command. This proposal biases the search toward temporally coherent directions supported by the collected actions, reducing---though not eliminating---the world-model exploitations that arise from unconstrained action search.

When the monitor triggers, we query $\pibase$ once for its nominal action chunk $\action^{\pibase}_{t:t+\ph-1}$ and initialize the CEM population around it using Eq.~\eqref{eq:pca_noise}. Each candidate is rolled out through $\wm$ and scored by Eq.~\eqref{eqn:cost_function}; CEM then refits the proposal to the top-scoring elites and repeats for a fixed number of iterations. The planner executes the first $\ch\leq\ph$ actions of the best sequence and returns control to the monitor, which re-invokes it if $\pibase$ has still not recovered.

\section{Experimental Results}\label{sec:result}

\subsection{Experimental Setup}
We evaluate our method on the ten-task LIBERO-Goal~\cite{libero} suite, using
$\pi_{0.5}$~\cite{pi05}'s LIBERO checkpoint as the nominal policy $\pibase$ throughout.
All ten tasks share the same kitchen tabletop scene with a Franka Panda manipulator and differ only in their language-based
task goals.
For detailed comparisons against computationally intensive baselines, we use a four-task representative subset --- Tasks t0 (\textit{open the middle drawer}), t3 (\textit{open the top drawer and put the bowl inside}), t4 (\textit{put the bowl on the top of the cabinet}) and t9 (\textit{put the wine bottle on the rack}), spanning $\pibase$'s competence range and covering drawer opening, pick-and-place, and a multi-step task.

The $\pi_{0.5}$-LIBERO checkpoint reaches $98.0\%$ success on LIBERO-Goal under the default benchmark setup, as reported by the official \texttt{openpi} implementation~\cite{openpi_repo}. Because this near-saturated performance provides few failures on which to evaluate intervention, we construct a perturbed evaluation setting by varying the initial manipulator configuration, which also reorients the wrist camera, and the initial $x,y$ locations of objects in the scene. We tune the perturbation magnitudes by task difficulty: up to $\pm0.1\chi$ radians on joints 4 and 5, $\pm0.2\pi\chi$ radians on the wrist roll joint, and $[x\pm\Upsilon,y\pm\Upsilon]$ meters for object locations. Under this protocol, $\pibase$ achieves an average success rate of $72.6\%$ (Table~\ref{tab:suite}). These perturbations define the deployment distribution, and training rollouts and test episodes use the same perturbation parameters.

For each task, we collect 200 rollouts of $\pibase$ under the same perturbation parameters used at test time, each capped at 300 steps, and labeled success or failure (timeout) by the simulator. We train the liveness value function $\vf$ and world model $\wm$ on these rollouts on a single NVIDIA GeForce RTX 4090 GPU. We additionally supplement the world model $\wm$ with $53$ minutes of the LIBERO-Goal teleoperated demonstrations used as play data, pooled
across all ten tasks.
For any given task, $90\%$ of these demonstrations correspond to other tasks and therefore provide primarily image--action coverage rather than demonstrations of task-correct behavior.This additional play data need not come from expert demonstrations: trained on the $200$ policy rollouts alone, our method still gains $14.7$pp on Task 0 (Sec.~\ref{sec:ablations}).

We run 300 paired episodes per task, where all methods receive the same initial condition, and $\pi_{0.5}$'s sampling is fixed to the same seed across methods. Without intervention, the methods therefore reproduce the nominal policy's trajectory exactly; trajectories diverge only when a monitor triggers, attributing differences in outcomes to the interventions alone.

\begin{table}[t]
\centering
\caption{\textbf{Full LIBERO-Goal suite.} Success rate (\%), $n\!=\!300$ paired
trials per task. Rescued and Broken count interventions that flip a trial's outcome; $\Delta$ is their difference.}\label{tab:suite}
\footnotesize
\setlength{\tabcolsep}{1.5pt}
\begin{tabular}{@{}l *{10}{c} @{\hspace{3pt}}c@{}}
\toprule
 & t0 & t1 & t2 & t3 & t4 & t5 & t6 & t7 & t8 & t9 & Mean \\
\midrule
$\chi$ & $1.15$ & $2.00$ & $1.40$ & $1.15$ & $2.00$ & $1.40$ & $1.40$ & $2.00$ & $1.85$ & $1.10$ & \\
$\Upsilon$ & $.02$ & $.10$ & $.04$ & $.00$ & $.10$ & $.04$ & $.04$ & $.10$ & $.10$ & $.02$ & \\
\midrule
$\pibase$ & $66.7$ & $81.0$ & $\mathbf{80.3}$ & $57.0$ & $84.3$ & $58.0$ & $76.3$ & $98.0$ & $71.7$ & $53.0$ & $72.6$ \\
Ours & $\mathbf{93.7}$ & $\mathbf{84.7}$ & $79.3$ & $\mathbf{67.0}$ & $\mathbf{89.3}$ & $\mathbf{70.0}$ & $\mathbf{80.7}$ & $\mathbf{98.3}$ & $\mathbf{78.7}$ & $\mathbf{73.7}$ & $\mathbf{81.5}$ \\
\midrule
Rescued & $86$ & $13$ & $9$ & $37$ & $17$ & $58$ & $23$ & $1$ & $37$ & $88$ & $\mathbf{369}$ \\
Broken & $5$ & $2$ & $12$ & $7$ & $2$ & $22$ & $10$ & $0$ & $16$ & $26$ & $\mathbf{102}$ \\
\midrule
$\Delta$ & $27.0$ & $3.7$ & $-1.0$ & $10.0$ & $5.0$ & $12.0$ & $4.3$ & $0.3$ & $7.0$ & $20.7$ & $\mathbf{8.9}$ \\
\bottomrule
\end{tabular}\vspace{-12pt}
\end{table}

\subsection{Deployment-time Improvement Across the Suite}

Across all ten tasks and $3{,}000$ paired trials, our method improves $\pibase$ from
$72.6\%$ to $81.5\%$ ($+8.9$pp; Table~\ref{tab:suite}). The monitor reliably identifies executions requiring intervention: it fires on $94\%$ of the trials that $\pibase$ would have failed and on only $16\%$ of those that $\pibase$ would have succeeded. On the other hand, intervening on trials that $\pibase$ would otherwise have completed creates an opportunity to disrupt successful behavior. Across the suite, the filter rescues $369$ nominal-policy failures but breaks $102$ nominal-policy successes, for a net gain of $267$ successful trials. This rescue--breakage tradeoff is a core tension when filtering with learned models.

\begin{figure*}[t]
 \centering
3 \includegraphics[width=0.86\textwidth]{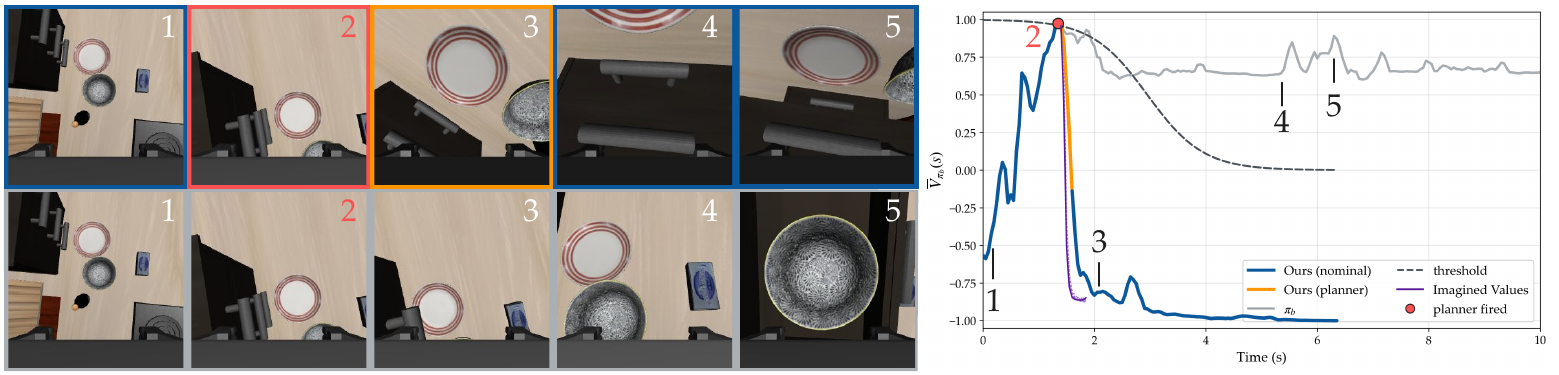}
 \caption{A sample visualization of the wrist camera in Task 0 (\textit{open the middle drawer}). The proposed method (top) guides the robot towards the correct middle drawer, while the nominal policy (bottom) erroneously targets the top drawer.} \vspace{-12pt}
 \label{fig:libero_t0_viz}
\end{figure*}

The monitor fires in proportion to the competency of $\pibase$ on that task. On Task 7 (\textit{turn on the stove}), where $\pibase$ already succeeds $98.0\%$ of the time, it fires on $5$ of $300$ trials and nets one additional success. On Task 9, the hardest task, it fires on $211$ and nets 62 additional successes. 
Task 0 shows the largest gain ($+27.0$pp), where the task's progression is visually legible to $\vfmean$ through the configuration of the cabinet drawers, and $\wm$ models the relevant drawer dynamics sufficiently well for planning. Here, $\pibase$ often times out after opening the wrong drawer and getting stuck; our method instead plans toward the correct drawer.
Fig.~\ref{fig:libero_t0_viz} illustrates one such Task~0 episode.
The nominal policy begins by moving correctly toward the cabinet but then reaches for the handle of the top drawer.
The monitor registers the drift as a threshold exceedance, and the planner steers the end
effector away from the top drawer and toward the lower drawers. $\pibase$ then resumes control,
selects the middle drawer, and opens it. 
The two trajectories are identical until the monitor triggers. Without intervention, the policy not only opens the wrong drawer but continues executing a different task, placing the bowl inside it. This failure involves neither an unsafe state nor a local control error, legible only as a loss of progress toward the commanded goal.

The aggregate improvement on the task suite is limited by both ceiling effects and occasional harmful interventions. Task~7 is already near ceiling ($98.0\!\rightarrow\!98.3$), leaving little room for improvement. Task~2 (\textit{put the wine bottle on top of the cabinet}) is the only task on which performance regresses ($-1.0$pp). Here, $\wm$ hallucinates grasps on the thin wine bottle, sometimes causing the robot to tip over the bottle. In contrast, Task~9 presents more recoverable failures involving the same object and improves by $+20.7$pp.

\begin{table}[t]
\centering
\caption{\textbf{Comparison against direct-improvement and filtering methods.} Success
rate (\%), $n\!=\!300$ per task. TD3+BC and IQL \emph{replace} $\pibase$; Filtered-BC
\emph{retrains} it with white-box weight access; the remaining methods \textit{filter} it.}
\label{tab:baselines}
\footnotesize
\setlength{\tabcolsep}{3pt}
\renewcommand{\arraystretch}{1.05}
\begin{tabular}{@{}l c c c c c c@{}}
\toprule
& & \multicolumn{4}{c}{Black-box access}
    & \multicolumn{1}{c}{White-box reference} \\
\cmidrule(lr){3-6} \cmidrule(lr){7-7}
Task & $\pibase$ & TD3+BC & IQL & LatentSafe & Ours & Filtered-BC \\
\midrule
t0 & $66.7$ & $56.3$ & $42.7$ & $66.3$ & $\mathbf{93.7}$ & $81.0$ \\
t3 & $57.0$ & $7.3$ & $34.3$ & $57.0$ & $67.0$ & $\mathbf{75.3}$ \\
t4 & $84.3$ & $60.0$ & $54.3$ & $84.7$ & $\mathbf{89.3}$ & $\mathbf{89.3}$ \\
t9 & $53.0$ & $35.0$ & $37.3$ & $28.7$ & $\mathbf{73.7}$ & $67.0$ \\
\midrule
Mean & $65.2$ & $39.6$ & $42.2$ & $59.2$ & $\mathbf{80.9}$ & $78.2$ \\
\bottomrule\vspace{-24pt}
\end{tabular}
\end{table}

\subsection{Comparison Against Other Policy Improvement Methods}
Table~\ref{tab:baselines} compares our method against policy-improvement approaches spanning two access regimes. Our primary comparisons are methods that, like ours, can be constructed using only rollout data and black-box policy queries. 
\textbf{TD3+BC}~\cite{fujimoto2021minimalist} and \textbf{IQL}~\cite{kostrikov2021offline} are offline-RL baselines that train replacement policies from the collected rollouts. 
In contrast, \textbf{LatentSafe}~\cite{latent_safety_filter} is a filtering baseline that adopts the same monitoring-and-intervention structure as our method. It learns an unsafe set and a safety-preserving policy via latent-space reachability analysis. We adapt it to our liveness (task-completion) formulation, using its learned policy as the fallback $\pif$ in place of our planner and our same sigmoid trigger schedule. We additionally compare against \textbf{Filtered-BC}: it fine-tunes all $3.3$B parameters of the $\pi_{0.5}$ checkpoint on the successful rollouts our monitor and world model see.
It therefore requires white-box access to the policy, its training pipeline, and H100-scale training compute. Because these requirements violate our assumed black-box access model, we treat Filtered-BC as a white-box reference rather than a directly competing baseline.

Among the black-box methods, ours achieves the highest mean success rate at $80.9\%$, compared with $59.2\%$ for LatentSafe, $42.2\%$ for IQL, and $39.6\%$ for TD3+BC. Although TD3+BC and IQL require no access to the parameters of $\pibase$, they discard the capabilities encoded in the original policy. Their performance is also limited by the narrow action coverage of the rollout data: $\pibase$ is nearly deterministic and rarely revisits the same state, providing little within-state action variation from which to infer better actions. Consequently, IQL's policy-extraction step approaches behavior cloning on potentially suboptimal rollouts. TD3+BC, which we augment with transitions generated by the learned world model, is additionally exposed to model error and performs particularly poorly on Task~3 ($7.3\%$), where accurately modeling the drawer state is necessary for the subsequent bowl placement. Similarly, LatentSafe rarely intervenes, and its interventions do not reliably recover progress. On Task~9, for example, it breaks $80$ nominal-policy successes while rescuing only $7$ failures, leaving $\pibase$ worse off overall ($59.2\%$ versus $65.2\%$ mean success). Its fallback is a liveness-preserving controller trained to keep the system within the set of states from which the goal remains reachable. Such an objective is well suited to maintaining invariance, but provides limited guidance once task progress has already been lost: returning to the reachable set requires recovery, rather than simply remaining within it.

Finally, our black-box method achieves comparable mean performance to the white-box Filtered-BC baseline ($80.9\%$ versus $78.2\%$), without accessing or modifying the nominal policy's weights.
Filtering and fine-tuning are also complementary rather than mutually exclusive. For example, applying our filter on top of Filtered-BC, without additional data or further access to its weights, raises its mean success rate from $78.2\%$ to $79.8\%$. Task~0 improves from $81.0\%$ to $95.3\%$, whereas Task~3 decreases from $75.3\%$ to $66.0\%$. The Task~3 monitor remains similarly effective after fine-tuning, flagging $87.8\%$ of failures under Filtered-BC compared with $86.8\%$ under $\pibase$. The degradation therefore occurs downstream of detection, consistent with this task's world-model limitations.

\begin{table}[t]
\centering
\caption{\textbf{Perturbation-severity stress test.} 
Success rate (\%) when one perturbation magnitude is increased by $50\%$ relative to the main experimental setting and the other is held fixed. ``Set'' denotes the perturbed value.
$^{*}$Task 3 has no
nominal object perturbation, so its object condition introduces $\pm.02$ rather than scaling.}
\label{tab:ood}
\footnotesize
\setlength{\tabcolsep}{3pt}
\renewcommand{\arraystretch}{1.05}
\begin{tabular}{@{}l rrrr rrrr@{}}
\toprule
 & \multicolumn{4}{c}{$+50\%$ Config.\ perturbation $\chi$} & \multicolumn{4}{c}{$+50\%$ Object perturbation $\Upsilon$} \\
\cmidrule(lr){2-5} \cmidrule(lr){6-9}
Task & Set & $\pibase$ & Ours & $\Delta$ & Set & $\pibase$ & Ours & $\Delta$ \\
\midrule
t0 & $1.725$ & $43.0$ & $\mathbf{71.0}$ & $+28.0$ & $.03$ & $67.0$ & $\mathbf{89.0}$ & $+22.0$ \\
t3 & $1.725$ & $41.0$ & $\mathbf{48.0}$ & $+7.0$ & $^{*}.02$ & $54.0$ & $\mathbf{68.0}$ & $+14.0$ \\
t4 & $3.000$ & $57.0$ & $\mathbf{70.0}$ & $+13.0$ & $.15$ & $\mathbf{78.0}$ & $77.0$ & $-1.0$ \\
t9 & $1.650$ & $37.0$ & $\mathbf{61.0}$ & $+24.0$ & $.03$ & $51.0$ & $\mathbf{62.0}$ & $+11.0$ \\
\midrule
Mean & & $44.5$ & $\mathbf{62.5}$ & $+18.0$ & & $62.5$ & $\mathbf{74.0}$ & $+11.5$ \\
\bottomrule\vspace{-27pt}
\end{tabular}
\end{table}

\subsection{Robustness Analysis and Component Ablations}\label{sec:ablations}
\noindent \textbf{Perturbation-severity stress test.}
We conduct a limited stress test by increasing each perturbation magnitude by $50\%$, one at a time, leaving $\vf$ and $\wm$ unchanged ($n\!=\!100$ per condition; Table~\ref{tab:ood}). This experiment characterizes sensitivity to more severe perturbations along the same axes used in our main evaluation. Increasing the configuration perturbation degrades $\pibase$ most severely, to $44.5\%$, and yields our largest mean improvement ($+18.0$pp). Increasing the object perturbation yields an improvement of $+11.5$pp. Our method improves over $\pibase$ in seven of the eight conditions; the exception is Task~4 at the largest $\pm15$\,cm object displacement, where it underperforms $\pibase$ by $1.0$pp.

\vspace{0.3em}
\noindent \textbf{Component ablations.}
We next ablate the principal design choices in the monitor and planner (Table~\ref{tab:ablations}). Monitor ablations vary one component at a time ($n\!=\!100$ per task), averaged across the subset.
Removing the sigmoid threshold schedule and replacing it with a fixed
zero threshold reduces mean success by $8.5$pp; collapsing the five-member value ensemble to a single network costs $8.3$pp; and removing temporal smoothing costs $4.3$pp. The threshold schedule and value ensemble therefore have the largest effects among the monitor design choices, although even the weakest
variant still improves over $\pibase$ by $6.0$pp.

The liveness value is the component around which both monitoring and planning are organized, so we test it within the complete intervention pipeline. We replace it with the distributional Monte Carlo value function of \cite{pi06_star}, which discretizes trajectory progress into timestep classes and predicts a distribution over them, while leaving the world model, planner,
trigger schedule, and threshold unchanged. This substitution reduces mean success by $13.1$pp, the largest effect measured in our ablations. The two objectives carry different inductive biases: the liveness recursion propagates the non-goal value until goal attainment is observed, whereas the classifier tends toward frequently represented classes mid-trajectory. Such predictions can still detect failures well but provide a less effective planning objective. Consistent with this explanation, MC-D rescues $152$ nominal-policy failures, compared with $211$ for our liveness value, while breaking $120$ nominal-policy successes, compared with $38$ for ours.

On the planner side, replacing our learned action basis with iCEM-colored noise ($\beta$=2, Brownian)~\cite{pinneri2021sample} costs $4.7$ pp, demonstrating the efficacy of the proposed planning algorithm.
We also vary the replanning frequency, executing only $2$ of the planned actions rather than $5$ leads to a modest cost of $-1.3$pp. 

Finally, we ask how much of the result depends on privileged demonstration data. Refitting both the dynamics model and the planner's action basis on the $200$ policy rollouts alone --- removing play and demonstration data from every learned component --- still improves Task 0 from $66.7\%$ to $81.3\%$ ($+14.7$pp), retaining $54\%$ of the full gain and still exceeding our strongest baseline. We attribute the gap to state--action support rather than task-specific knowledge: without play data the rollout actions need $23$ principal components to capture $99\%$ of their variance against $15$ with it, so our fixed $13$-component basis spans materially less.

\begin{table}[t]
\centering
\caption{\textbf{Component ablations.} Success rate (\%) with one component
varied from the full configuration; all others pinned. $\Delta$ is the change in
mean success.}
\label{tab:ablations}
\footnotesize
\setlength{\tabcolsep}{4pt}
\renewcommand{\arraystretch}{1.05}
\begin{tabular}{@{}l cccc @{\hspace{4pt}}c c@{}}
\toprule
Configuration & t0 & t3 & t4 & t9 & Mean & $\Delta$ \\
\midrule
$\pibase$ (unfiltered) & $73.0$ & $56.0$ & $85.0$ & $57.0$ & $67.8$ & $-14.5$ \\
\midrule
\multicolumn{7}{@{}l}{\emph{Monitor design ablations} ($n\!=\!100$)} \\
Ours (full) & $\mathbf{96.0}$ & $\mathbf{66.0}$ & $89.0$ & $\mathbf{78.0}$ & $\mathbf{82.3}$ & --- \\
\quad no trigger schedule & $91.0$ & $50.0$ & $\mathbf{93.0}$ & $61.0$ & $73.8$ & $-8.5$ \\
\quad 1-member VF & $91.0$ & $58.0$ & $90.0$ & $57.0$ & $74.0$ & $-8.3$ \\
\quad 2-member VF & $\mathbf{96.0}$ & $62.0$ & $91.0$ & $62.0$ & $77.8$ & $-4.5$ \\
\quad no smoothing ($W\!=\!1$) & $90.0$ & $65.0$ & $89.0$ & $68.0$ & $78.0$ & $-4.3$ \\
\quad horizon $h\!=\!2$ & $95.0$ & $65.0$ & $89.0$ & $75.0$ & $81.0$ & $-1.3$ \\
\midrule
\multicolumn{7}{@{}l}{{\emph{Component substitutions} ($n\!=\!300$)}} \\
Ours (full) & $\mathbf{93.7}$ & $\mathbf{67.0}$ & $89.3$ & $\mathbf{73.7}$ & $\mathbf{80.9}$ & --- \\
\quad MC-D value signal & $79.3$ & $62.7$ & $84.0$ & $45.3$ & $67.8$ & $-13.1$ \\
\quad iCEM proposal & $88.3$ & $61.3$ & $\mathbf{90.0}$ & $65.0$ & $76.2$ & $-4.7$ \\
\bottomrule\vspace{-24pt}
\end{tabular}
\end{table}

\vspace{-3pt}
\subsection{Hardware Experiment} \label{sec:hardware}

We deploy our method on a lunchbox-opening task (Fig.~\ref{fig:header}) using a Franka Panda 7-DoF manipulator and a visuomotor diffusion policy~\cite{diffusion_policy}, with inference performed on an RTX 5090 GPU. The nominal policy $\pibase$ is trained on 105 expert demonstrations to grasp the lip of a deformable cloth lid and open the lunchbox without spilling its contents, in under 30 seconds; all collected at the same nominal initial configuration.

At test time, we perturb the initial manipulator configuration by an angle in $[0,9]^\circ$ and use the same 30-second wall-clock limit. 
Under this test distribution, the nominal policy achieves a success rate of $40\%$ over 120 trials. 
Next, we use these 120 on-policy rollouts to train the liveness value $\vfmean$ and world model $\wm$. The world-model data are supplemented with the 105 expert demonstrations, 15 minutes of teleoperator practice, and 30 minutes of targeted play consisting of smooth motions around the lunchbox workspace.

Our method, run for $50$ episodes under the same conditions, recovers $60.0\%$ ($+20$pp). The improvement is not confined to small perturbations (Table~\ref{tab:hardware}). The largest observed difference occurs in the most difficult band, $[6.5,9]^\circ$, where the nominal policy succeeds only $22.7\%$, compared with $70.6\%$ with our filter in the loop. 

\begin{wraptable}{r}{0.5\columnwidth}
\centering
\caption{\textbf{Hardware lunchbox opening task.} Success rate (\%) for initial-configuration perturbations in $[0,9]^\circ$.}
\label{tab:hardware}
\footnotesize
\setlength{\tabcolsep}{2pt}
\renewcommand{\arraystretch}{1.05}
\begin{tabular}{@{}l r r r@{}}
\toprule
Perturb. & $\pibase$ & $+$Filt. & $\Delta$ \\
\midrule
all & $40.0$ & $\mathbf{60.0}$ & $\mathbf{+20.0}$ \\
\midrule
$0^\circ$ & $80.0$ & $100.0$ & $+20.0$ \\
$0$--$4^\circ$ & $54.2$ & $71.4$ & $+17.3$ \\
$4$--$6.5^\circ$ & $34.8$ & $43.5$ & $+8.7$ \\
$6.5$--$9^\circ$ & $22.7$ & $\mathbf{70.6}$ & $\mathbf{+47.9}$ \\
\bottomrule
\end{tabular}
\vspace{-6pt}
\end{wraptable}
We further decompose the 50 filtered trials according to whether the monitor intervened. The monitor remained inactive in 16 trials, all of which succeeded. It triggered at least once in the remaining 34 trials, of which 14 succeeded. Thus, every observed failure followed at least one monitor trigger and detection is not what limits the result. 

Fig.~\ref{fig:header}(\textbf{\textit{a}}) illustrates one such intervention. At $t_1 \approx 1\,\text{s}$, the monitor observes that $\pibase$ has erroneously closed the gripper without grasping the lid. The smoothed liveness value exceeds the intervention threshold, $\vfmean(s_{t_1})\!>\!\threshold_{t_1}$, and triggers the planner. 
The planning traces and decoded world-model predictions show the imagined futures of the best candidates whose actions are executed. Control alternates between $\pibase$ and $\pif$ across successive interventions until $t_4 \approx 6.5\,\text{s}$, when the planner predicts a value-minimizing plan for a complete grasp-and-open maneuver and the system executes it, after which $\pibase$ resumes to finish opening the lid.

\subsection{Deployment Cost}
A full planning cycle takes $108\!\pm\!3$\ms in addition to $180\!\pm\!3$\ms for a single
query to the nominal policy: intervening costs less than a single policy query, and far less than the $n$ queries per step that re-ranking approaches pay. 
Both inference costs are amortized over the four-step action chunk they produce, so planning adds $27$\ms per executed control step against the policy's own $45$\ms.
This corresponds to an effective control rate of approximately $13.9$\,Hz during intervention, while planning incurs no overhead when the monitor remains inactive.

Evaluating many candidate futures is affordable within this budget because the world model processes them in parallel. A single 16-step imagined rollout takes $9.5$\ms, while a batch of 256 such rollouts takes $44.2$\ms. This sublinear scaling with our compact $29.6$M-parameter world model and five-$1.85$M-parameter value ensemble makes sample-based planning viable within the control loop.
The additional inference nevertheless reduces the number of actions executed within the fixed $30\,\text{s}$ episode limit. Across the hardware evaluation, the filtered system executes $20\%$ fewer control steps than the unfiltered, yet still achieves a higher success rate (Table~\ref{tab:hardware}). 

For context, VLA-in-the-Loop~\cite{xu2026vlaintheloop} and DreamAvoid~\cite{fan2026dreamavoid} report approximately $20$\,s and $2$\,s per video-based intervention, respectively, while RoboMonkey~\cite{kwok2025robomonkey} operates at approximately $1.5$\,Hz when sampling and verifying multiple policy proposals. These use different policies and hardware, but indicate the computational scale of existing test-time correction approaches. In their respective reported settings, our $108$\ms planning cycle is approximately $6$--$185\times$ shorter, enabling corrective planning without pausing execution.

\section{Limitations and Conclusion}
\label{sec:conclusion}
We present a monitoring-and-intervention framework for improving visuomotor manipulation policies at deployment time using only black-box access. A policy-conditioned liveness value monitors progress and, when needed, guides a short-horizon planner toward states from which the nominal policy can recover; planning, evaluating, and executing within the real-time control loop. Across the LIBERO-Goal suite, our method improves the success rate of $\pi_{0.5}$ from $72.6\%$ to $81.5\%$, outperforming offline-RL and alternative filtering baselines. We further demonstrate the framework with a diffusion policy on a Franka Panda manipulator, where it improves success from $40\%$ to $60\%$.

The method does not, however, guarantee that every intervention improves the execution. Table~\ref{tab:suite} reveals two distinct failure mechanisms. First, an inaccurate liveness value can trigger on an execution that the nominal policy would have completed, creating an opportunity for the planner to disrupt an otherwise successful trajectory. Second, even when intervention is warranted, errors in the learned world model compound over autoregressive rollouts and can be exploited by the planner: an action sequence may score well under the imagined dynamics but fail when executed. Moreover, the framework requires labeled rollouts of the nominal policy and auxiliary transition data; it is not a zero-shot improvement method.
Finally, extending the framework to longer-horizon tasks, partially observable progress, and shared models that transfer across tasks remains an important direction.
	


\bibliographystyle{IEEEtran}
\bibliography{refs}

\end{document}

%% file: macros.tex
\newcommand{\task}{\mathcal{I}}
\newcommand{\pibase}{\pi_b}
\newcommand{\pif}{{\pi_{f}}}
\newcommand{\monitor}{\mathcal{M}}

\newcommand{\vf}{V_{\pibase}}
\newcommand{\vfmean}{\overline{V}_{\pibase}}
\newcommand{\threshold}{\eta}
\newcommand{\lengthcurr}{\kappa}

\newcommand{\wm}{\phi}
\newcommand{\state}{s}
\newcommand{\imgstate}{\hat{\state}}
\newcommand{\action}{a}

\newcommand{\cost}{J}
\newcommand{\ph}{H}
\newcommand{\ch}{h}

\newcommand{\vfred}{\textcolor[HTML]{FA5252}{\vf}}
\newcommand{\hwnote}[1]{\textcolor{blue}{HW:#1}}
\newcommand{\ccnote}[1]{\textcolor{red}{CC: #1}}

\newcommand{\pending}{\textcolor{red}{–}}
\newcommand{\dr}[1]{{\color{orange}#1}}
\newcommand{\drnote}[1]{{\color{orange}\textbf{[#1]}}}
\newcommand{\cut}[1]{{\color{red}#1}}
\newcommand{\ms}{\,ms\xspace}

%% file: refs.bib
@INPROCEEDINGS{latent_safety_filter, 
    AUTHOR    = {Kensuke Nakamura AND Lasse Peters AND Andrea Bajcsy}, 
    TITLE     = {{Generalizing Safety Beyond Collision-Avoidance via Latent-Space Reachability Analysis}}, 
    BOOKTITLE = {Proceedings of Robotics: Science and Systems}, 
    YEAR      = {2025}, 
    ADDRESS   = {LosAngeles, CA, USA}, 
    MONTH     = {June}, 
    DOI       = {10.15607/RSS.2025.XXI.113} 
}

@misc{dino_wm,
	title = {{DINO}-{WM}: {World} {Models} on {Pre}-trained {Visual} {Features} enable {Zero}-shot {Planning}},
	shorttitle = {{DINO}-{WM}},
	url = {http://arxiv.org/abs/2411.04983},
	doi = {10.48550/arXiv.2411.04983},
	urldate = {2025-06-03},
	publisher = {arXiv},
	author = {Zhou, Gaoyue and Pan, Hengkai and LeCun, Yann and Pinto, Lerrel},
	month = feb,
	year = {2025},
	note = {arXiv:2411.04983 [cs]},
}

@article{pi05,
  title={{$\pi_{0.5}$}: a Vision-Language-Action Model with Open-World Generalization},
  author={Intelligence, Physical and Black, Kevin and Brown, Noah and Darpinian, James and Dhabalia, Karan and Driess, Danny and Esmail, Adnan and Equi, Michael and Finn, Chelsea and Fusai, Niccolo and others},
  journal={arXiv preprint arXiv:2504.16054},
  year={2025}
}

@article{pi06_star,
  title={{$\pi_{0.6}^*$}: a VLA That Learns From Experience},
  author={Intelligence, Physical and Amin, Ali and Aniceto, Raichelle and Balakrishna, Ashwin and Black, Kevin and Conley, Ken and Connors, Grace and Darpinian, James and Dhabalia, Karan and DiCarlo, Jared and others},
  journal={arXiv preprint arXiv:2511.14759},
  year={2025}
}

@article{gemini_robotics_1_5,
  title={Gemini robotics 1.5: Pushing the frontier of generalist robots with advanced embodied reasoning, thinking, and motion transfer},
  author={Team, Gemini Robotics and Abdolmaleki, Abbas and Abeyruwan, Saminda and Ainslie, Joshua and Alayrac, Jean-Baptiste and Arenas, Montserrat Gonzalez and Balakrishna, Ashwin and Batchelor, Nathan and Bewley, Alex and Bingham, Jeff and others},
  journal={arXiv preprint arXiv:2510.03342},
  year={2025},
  url={https://arxiv.org/abs/2510.03342}
}

@article{open_vla,
  title={Openvla: An open-source vision-language-action model},
  author={Kim, Moo Jin and Pertsch, Karl and Karamcheti, Siddharth and Xiao, Ted and Balakrishna, Ashwin and Nair, Suraj and Rafailov, Rafael and Foster, Ethan and Lam, Grace and Sanketi, Pannag and others},
  journal={arXiv preprint arXiv:2406.09246},
  year={2024}
}

@misc{diffusion_policy,
	title = {Diffusion {Policy}: {Visuomotor} {Policy} {Learning} via {Action} {Diffusion}},
	shorttitle = {Diffusion {Policy}},
	url = {http://arxiv.org/abs/2303.04137},
	doi = {10.48550/arXiv.2303.04137},
	urldate = {2026-03-18},
	publisher = {arXiv},
	author = {Chi, Cheng and Feng, Siyuan and Du, Yilun and Xu, Zhenjia and Cousineau, Eric and Burchfiel, Benjamin and Song, Shuran},
	month = jun,
	year = {2023},
	note = {arXiv:2303.04137 [cs]
version: 4},
}

@misc{latent_policy_barrier,
	title = {Latent {Policy} {Barrier}: {Learning} {Robust} {Visuomotor} {Policies} by {Staying} {In}-{Distribution}},
	shorttitle = {Latent {Policy} {Barrier}},
	url = {http://arxiv.org/abs/2508.05941},
	doi = {10.48550/arXiv.2508.05941},
	urldate = {2026-04-20},
	publisher = {arXiv},
	author = {Sun, Zhanyi and Song, Shuran},
	month = aug,
	year = {2025},
	note = {arXiv:2508.05941 [cs]},
}

@misc{bansalHamiltonJacobiReachabilityBrief2017b,
	title = {Hamilton-{Jacobi} {Reachability}: {A} {Brief} {Overview} and {Recent} {Advances}},
	shorttitle = {Hamilton-{Jacobi} {Reachability}},
	url = {http://arxiv.org/abs/1709.07523},
	doi = {10.48550/arXiv.1709.07523},
	urldate = {2026-04-27},
	publisher = {arXiv},
	author = {Bansal, Somil and Chen, Mo and Herbert, Sylvia and Tomlin, Claire J.},
	month = sep,
	year = {2017},
	note = {arXiv:1709.07523 [eess]},
}

@article{wang2026_policy_eval,
  title={Offline Policy Evaluation for Manipulation Policies via Discounted Liveness Formulation},
  author={Wang, Hao and Bowden, Joshua and Crosby, Colton and Bansal, Somil},
  journal={arXiv preprint arXiv:2605.11479},
  year={2026}
}

@misc{hafnerMasteringDiverseDomains2024,
	title = {Mastering {Diverse} {Domains} through {World} {Models}},
	url = {http://arxiv.org/abs/2301.04104},
	doi = {10.48550/arXiv.2301.04104},
	urldate = {2026-04-22},
	publisher = {arXiv},
	author = {Hafner, Danijar and Pasukonis, Jurgis and Ba, Jimmy and Lillicrap, Timothy},
	month = apr,
	year = {2024},
	note = {arXiv:2301.04104 [cs]},
}

@inproceedings{learning_latent_plans_from_play,
  title={Learning latent plans from play},
  author={Lynch, Corey and Khansari, Mohi and Xiao, Ted and Kumar, Vikash and Tompson, Jonathan and Levine, Sergey and Sermanet, Pierre},
  booktitle={Conference on robot learning},
  pages={1113--1132},
  year={2020},
  organization={Pmlr}
}

@article{borquez2024safety,
  title={On safety and liveness filtering using Hamilton--Jacobi reachability analysis},
  author={Borquez, Javier and Chakraborty, Kaustav and Wang, Hao and Bansal, Somil},
  journal={IEEE Transactions on Robotics},
  volume={40},
  pages={4235--4251},
  year={2024},
  publisher={IEEE}
}

@article{siglip2,
  title={Siglip 2: Multilingual vision-language encoders with improved semantic understanding, localization, and dense features},
  author={Tschannen, Michael and Gritsenko, Alexey and Wang, Xiao and Naeem, Muhammad Ferjad and Alabdulmohsin, Ibrahim and Parthasarathy, Nikhil and Evans, Talfan and Beyer, Lucas and Xia, Ye and Mustafa, Basil and others},
  journal={arXiv preprint arXiv:2502.14786},
  year={2025},
  url={https://arxiv.org/abs/2502.14786}
}

@article{droid_dataset,
  title={Droid: A large-scale in-the-wild robot manipulation dataset},
  author={Khazatsky, Alexander and Pertsch, Karl and Nair, Suraj and Balakrishna, Ashwin and Dasari, Sudeep and Karamcheti, Siddharth and Nasiriany, Soroush and Srirama, Mohan Kumar and Chen, Lawrence Yunliang and Ellis, Kirsty and others},
  journal={arXiv preprint arXiv:2403.12945},
  year={2024},
  url={https://arxiv.org/abs/2403.12945}
}

@inproceedings{oxe_dataset,
  title={Open x-embodiment: Robotic learning datasets and rt-x models: Open x-embodiment collaboration 0},
  author={O’Neill, Abby and Rehman, Abdul and Maddukuri, Abhiram and Gupta, Abhishek and Padalkar, Abhishek and Lee, Abraham and Pooley, Acorn and Gupta, Agrim and Mandlekar, Ajay and Jain, Ajinkya and others},
  booktitle={2024 IEEE International Conference on Robotics and Automation (ICRA)},
  pages={6892--6903},
  year={2024},
  organization={IEEE}
}

@article{ames_2016_cbf_qp,
  title={Control barrier function based quadratic programs for safety critical systems},
  author={Ames, Aaron D and Xu, Xiangru and Grizzle, Jessy W and Tabuada, Paulo},
  journal={IEEE Transactions on Automatic Control},
  volume={62},
  number={8},
  pages={3861--3876},
  year={2016},
  publisher={IEEE}
}

@article{wang_2024_coop_safe_perf,
  title={\href{https://arxiv.org/abs/2409.06696}{Cooptimizing Safety and Performance With a Control-Constrained Formulation}},
  author={Wang, Hao and Dhande, Adityaya and Bansal, Somil},
  journal={IEEE Control Systems Letters},
  year={2024},
  publisher={IEEE}
}

@article{libero,
  title={Libero: Benchmarking knowledge transfer for lifelong robot learning},
  author={Liu, Bo and Zhu, Yifeng and Gao, Chongkai and Feng, Yihao and Liu, Qiang and Zhu, Yuke and Stone, Peter},
  journal={Advances in Neural Information Processing Systems},
  volume={36},
  pages={44776--44791},
  year={2023},
  url={https://arxiv.org/abs/2306.03310}
}

@misc{wm_play_chen2025spa,
      title={Internalizing World Models via Self-Play Finetuning for Agentic RL}, 
      author={Shiqi Chen and Tongyao Zhu and Zian Wang and Jinghan Zhang and Kangrui Wang and Siyang Gao and Teng Xiao and Yee Whye Teh and Junxian He and Manling Li},
      year={2025},
      eprint={2510.15047},
      archivePrefix={arXiv},
      primaryClass={cs.LG},
      url={https://arxiv.org/abs/2510.15047}, 
}

@inproceedings{pinneri2021sample,
  title={Sample-efficient cross-entropy method for real-time planning},
  author={Pinneri, Cristina and Sawant, Shambhuraj and Blaes, Sebastian and Achterhold, Jan and Stueckler, Joerg and Rolinek, Michal and Martius, Georg},
  booktitle={Conference on Robot Learning},
  pages={1049--1065},
  year={2021},
  organization={PMLR}
}

@article{fan2026dreamavoid,
  title={DreamAvoid: Critical-Phase Test-Time Dreaming to Avoid Failures in VLA Policies},
  author={Fan, Xianzhe and Lu, Yuxiang and Gao, Shenyuan and Wu, Xiaoyang and Han, Ruihua and Li, Manling and Zhao, Hengshuang},
  journal={arXiv preprint arXiv:2605.11750},
  year={2026}
}

@misc{xu2026vlaintheloop,
title={VLA-IN-THE-LOOP: Online Policy Correction with World Models for Robust Robotic Grasping},
author={Shaoqing Xu and Fang Li and Zhixiang Duan and Yifan Yang and Tianshi Xie and Zhi-Xin Yang},
year={2026},
url={https://openreview.net/forum?id=aT4LG8c6DE}
}

@article{liu2026vls,
  title={VLS: Steering Pretrained Robot Policies via Vision-Language Models},
  author={Liu, Shuo and Singh, Ishneet Sukhvinder and Xu, Yiqing and Duan, Jiafei and Krishna, Ranjay},
  journal={arXiv preprint arXiv:2602.03973},
  year={2026}
}

@article{kwok2025robomonkey,
  title={Robomonkey: Scaling test-time sampling and verification for vision-language-action models},
  author={Kwok, Jacky and Agia, Christopher and Sinha, Rohan and Foutter, Matt and Li, Shulu and Stoica, Ion and Mirhoseini, Azalia and Pavone, Marco},
  journal={arXiv preprint arXiv:2506.17811},
  year={2025}
}

@article{nakamoto2024_steer_generalist,
  title={Steering your generalists: Improving robotic foundation models via value guidance},
  author={Nakamoto, Mitsuhiko and Mees, Oier and Kumar, Aviral and Levine, Sergey},
  journal={arXiv preprint arXiv:2410.13816},
  year={2024}
}

@article{wagenmaker2025_steer_dp_latent_rl,
  title={Steering your diffusion policy with latent space reinforcement learning},
  author={Wagenmaker, Andrew and Nakamoto, Mitsuhiko and Zhang, Yunchu and Park, Seohong and Yagoub, Waleed and Nagabandi, Anusha and Gupta, Abhishek and Levine, Sergey},
  journal={arXiv preprint arXiv:2506.15799},
  year={2025}
}

@article{wang2025latent,
  title={Latent policy steering with embodiment-agnostic pretrained world models},
  author={Wang, Yiqi and Verghese, Mrinal and Schneider, Jeff},
  journal={arXiv preprint arXiv:2507.13340},
  year={2025}
}

@article{kostrikov2021offline,
  title={Offline reinforcement learning with implicit q-learning},
  author={Kostrikov, Ilya and Nair, Ashvin and Levine, Sergey},
  journal={arXiv preprint arXiv:2110.06169},
  year={2021}
}

@article{fujimoto2021minimalist,
  title={A minimalist approach to offline reinforcement learning},
  author={Fujimoto, Scott and Gu, Shixiang Shane},
  journal={Advances in neural information processing systems},
  volume={34},
  pages={20132--20145},
  year={2021}
}

@misc{openpi_repo,
  title={{openpi}: Open-Source Models and Packages for Robotics},
  author={{Physical Intelligence}},
  year={2025},
  note={{LIBERO} benchmark results reported for the \texttt{pi05\_libero} checkpoint},
  howpublished={\url{https://github.com/Physical-Intelligence/openpi/blob/main/examples/libero/README.md}}
}
